\documentclass[letterpaper]{article} 
\usepackage{aaai2026}  
\usepackage{times}  
\usepackage{helvet}  
\usepackage{courier}  
\usepackage[hyphens]{url}  
\usepackage{graphicx} 
\usepackage{natbib}  
\usepackage{caption} 
\usepackage{algorithm}
\usepackage{algorithmic}

\usepackage{newfloat}
\usepackage{listings}
\DeclareCaptionStyle{ruled}{labelfont=normalfont,labelsep=colon,strut=off} 
\floatstyle{ruled}
\newfloat{listing}{tb}{lst}{}
\floatname{listing}{Listing}
\nocopyright 

\usepackage{booktabs}
\usepackage{multirow}
\usepackage{amsmath}
\usepackage[misc]{ifsym}

\usepackage{colortbl} 

\definecolor{cellgray}{RGB}{200, 200, 200} 

\title{\ours{}: Enhancing Multimodal Misinformation Detection\\with Diverse, Factual, and Relevant Rationales}
\author{
    Herun Wan\textsuperscript{\rm 1}, Jiaying Wu\textsuperscript{\rm 2}, Minnan Luo\textsuperscript{\Letter\ \rm 1}, Xiangzheng Kong\textsuperscript{\rm 1}, Zihan Ma\textsuperscript{\rm 1}, Zhi Zeng\textsuperscript{\rm 1}
}
\affiliations{
    \textsuperscript{\rm 1}Xi'an Jiaotong University, \textsuperscript{\rm 2}National University of Singapore\\
    \texttt{wanherun@stu.xjtu.edu.cn}, \texttt{minnluo@xjtu.edu.cn}

}

\usepackage{bibentry}

\newcommand{\ours}[1]{\textsc{DiFaR}}

\begin{document}

\maketitle

\begin{abstract}
Generating textual rationales from large vision-language models (LVLMs) to support trainable multimodal misinformation detectors has emerged as a promising paradigm. However, its effectiveness is fundamentally limited by three core challenges: (\romannumeral 1) insufficient diversity in generated rationales, (\romannumeral 2) factual inaccuracies due to hallucinations, and (\romannumeral 3) irrelevant or conflicting content that introduces noise. We introduce \ours{}, a detector-agnostic framework that produces diverse, factual, and relevant rationales to enhance misinformation detection. \ours{} employs five chain-of-thought prompts to elicit varied reasoning traces from LVLMs and incorporates a lightweight post-hoc filtering module to select rationale sentences based on sentence-level factuality and relevance scores. Extensive experiments on four popular benchmarks demonstrate that \ours{} outperforms four baseline categories by up to 5.9\% and boosts existing detectors by as much as 8.7\%. Both automatic metrics and human evaluations confirm that \ours{} significantly improves rationale quality across all three dimensions.\footnote{Available at https://github.com/whr000001/DiFaR.}
 
\end{abstract}


\section{Introduction}

Large vision-language models (LVLMs) have achieved remarkable performance across a wide array of multimodal tasks, driven by their powerful reasoning and representation capabilities. However, their effectiveness remains limited in multimodal misinformation detection (MMD), a task that demands precise factual grounding and fine-grained, task-specific reasoning~\citep{DBLP:conf/iclr/Liu0LHXCHDH25, li2025cmie}. 

To harness the potential of LVLMs for identifying multimodal misinformation, recent work has proposed a collaborative paradigm that combines LVLMs with trainable detectors~\citep{zheng2025predictions}, which we refer to as the \textbf{LVLM-as-Enhancer}. In this framework, LVLMs are prompted to generate textual rationales (i.e., interpretable justifications or explanations), which are then paired with the original news article and passed to a downstream trainable detector. This design aims to harness the generalization strength of LVLMs while preserving the adaptability of task-specific models.

\begin{figure}[t]
    \centering
    \includegraphics[width=0.8\linewidth]{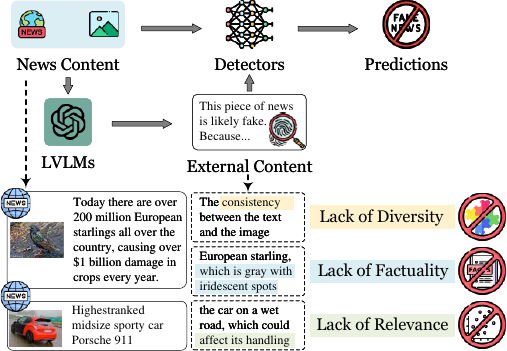}
    \caption{Illustration of three key challenges, namely, diversity, factuality, and relevance, in the LVLM-as-Enhancer paradigm for MMD, where LVLMs are prompted to generate explanatory rationales to support downstream detectors.}
    \label{fig: teaser}
\end{figure}

While this paradigm has shown early promise~\citep{tahmasebi2024multimodal, hu2024bad}, we identify three core limitations (illustrated in Figure~\ref{fig: teaser}) that hinder its full potential:

\begin{itemize}
    \item \textbf{Limited Diversity.} Most existing works concentrate on architectural innovations within the trainable detector~\citep{wang2024explainable}, paying limited attention to the quality and variation of generated rationales. These methods typically rely on a fixed prompt, which restricts the range of perspectives captured. As a result, they fail to exploit diverse reasoning signals that could enrich the interpretation of news content~\citep{wan2024dell}.
    \item \textbf{Limited Factuality.} LVLMs are prone to hallucinations~\citep{ji2023survey} and often generate content that deviates from verified facts~\citep{mallen2023not}. Consequently, the resulting rationales may introduce factual inaccuracies, which degrade the reliability of downstream detectors~\citep{pan2023risk}.
    \item \textbf{Limited Relevance.} Generated rationales frequently include loosely related or off-topic information, which may dilute or even conflict with the original article's claims~\citep{zheng2025predictions, xu2024knowledge}. This misalignment reduces the utility of the explanations and can compromise veracity assessment.
\end{itemize}

To address these limitations, we propose \ours{}, an MMD framework designed to produce rationales that are diverse, factual, and relevant. \ours{} is compatible with any existing trainable detector and operates without requiring architectural modifications. To enhance diversity, \ours{} incorporates multiple rationales derived from a set of five chain-of-thought (CoT) prompts, each targeting different aspects of the content, including textual details, visual features, and cross-modal consistency. This multi-prompt strategy allows for richer and more subtle reasoning.

To further improve rationale quality, \ours{} incorporates a post-hoc refinement module that filters individual rationale sentences based on factuality and relevance. For factuality, the module retrieves evidence from structured knowledge bases such as Wikipedia and compares it with the generated content~\citep{min2023factscore}. For relevance, it computes semantic similarity between the rationale and the source article using representation-based metrics~\citep{lewis2020retrieval}. Sentences with low scores are pruned, resulting in a distilled and trustworthy rationale set. 

We conduct extensive experiments on four multimodal misinformation datasets, covering both human-written and machine-generated news articles. Across all datasets, \ours{} consistently outperforms four representative categories of strong baselines, with up to 5.9\% relative accuracy improvement. Additionally, integrating \ours{} into existing detectors yields performance gains of up to 8.7\%. Ablation studies confirm that each component of \ours{} contributes meaningfully to its effectiveness. Further analysis, including human evaluation, validates that \ours{} improves the diversity, factuality, and relevance of the generated rationales.

\begin{figure*}[t]
    \centering
    \includegraphics[width=0.8\linewidth]{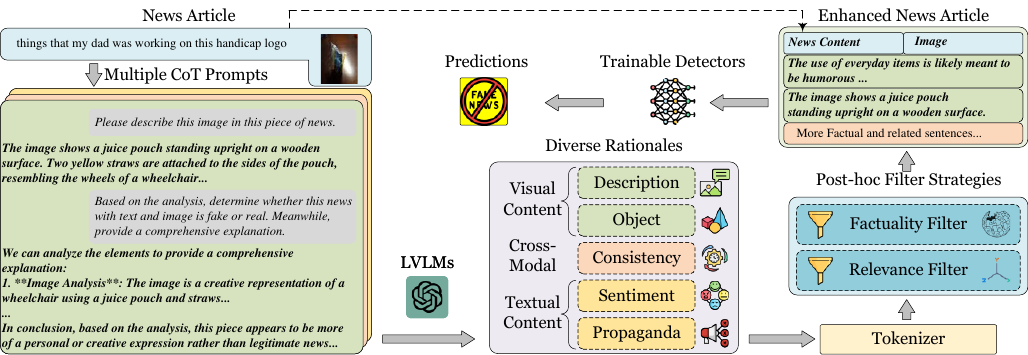}
    \caption{Overview of \ours{}. The framework integrates a simple yet effective structure that supports rationales of an arbitrary number and length. It employs five chain-of-thought prompts to promote reasoning diversity and two post-hoc refinement strategies to ensure factuality and relevance.}
    \label{fig: overview}
\end{figure*}

\section{Methodology}
\subsection{Preliminaries}

We consider the task of MMD as a binary classification problem. Each news instance consists of a textual component and a visual image, and the goal is to determine whether the news is real of fake. Formally, let $\mathcal{D}_{\textit{train}} = \{(T_i, V_i, y_i)\}_{i=1}^{N_{\textit{train}}}$ denote a training dataset of $N_{\textit{train}}$ labeled news articles, where $T_i$ is the text, $V_i$ is the associated image, and $y_i \in \{0,1\}$ is the ground-truth label. A trainable detector $f$ with parameters $\theta$ is trained to model the conditional distribution $p(y \mid T, V; f, \theta)$, with the objective of maximizing predictive accuracy on a test set $\mathcal{D}_{\textit{test}} = \{(T_i, V_i, y_i)\}_{i=1}^{N_{\textit{test}}}$.

Given a specific instance $(T, V, y)$ (we omit the index for clarity), conventional trainable detectors~\citep{chen2022cross, wang2023cross} first encode the modalities using frozen pre-trained encoders, yielding unimodal representations $\mathbf{t}$ (text) and $\mathbf{v}$ (image). These are fused into a joint representation $\mathbf{h}$ via a modality interaction module. The final prediction is computed as $p(y \mid T, V; f, \theta) \propto \exp(\mathrm{MLP}(\mathbf{h}))$, where $\mathrm{MLP}(\cdot)$ is a multi-layer perceptron. The predicted label is given by $\arg \max_y p(y \mid T, V; f, \theta)$.

The \emph{LVLM-as-Enhancer} paradigm extends this setup by leveraging an LVLM $\mathcal{G}$ to generate explanatory rationales $R = \mathcal{G}(T, V)$ for the input instance. These rationales are then encoded into $\mathbf{r}$ and incorporated into the detection pipeline through a specialized architecture that computes an enhanced representation $\mathbf{h}$. While this approach has shown early success~\citep{hu2024bad}, the over-engineered integration strategies utilized by existing efforts may limit generalizability to diverse rationale types. For example, EFND~\citep{wang2024explainable} introduces a structured module tailored to reason over debates, which may not generalize to alternative rationale formats such as sentiment-based reasoning about news veracity~\citep{zhang2021mining}.

\subsection{\ours{} Framework}
Figure~\ref{fig: overview} illustrates the overall architecture of \ours{}, a framework designed to robustly integrate multiple rationales into trainable multimodal misinformation detectors while preserving generality and scalability.

To improve compatibility with diverse rationales, \ours{} preserves the general pipeline of trainable detectors without introducing task-specific architectural changes. Given a news article $(T, V, y)$ and a set of $M$ LVLM-generated rationales $\{R_j\}_{j=1}^{M}$, we first concatenate all rationales with the original textual input to form an augmented input $\tilde{T} = [T; R_1; \dots; R_M]$, which is then passed to the detector $f$.

While this approach is intuitive and detector-agnostic, it presents two key challenges:
\begin{itemize}
    \item \textbf{Input length constraints.} Many detectors, such as those based on CLIP~\citep{radford2021learning}, have strict token limits (e.g., 77 tokens), making them unable to accommodate long concatenated inputs~\citep{chen2022cross}.
    \item \textbf{Ordering sensitivity.} The effectiveness of concatenated rationales may depend heavily on their order. Prior work has shown that sequence ordering significantly affects in-context learning performance~\citep{shi2024context}.  Exhaustively searching all permutations is computationally infeasible and unlikely to yield a universally optimal order. \end{itemize}

To overcome these limitations, \ours{} augments the textual modality at the representation level. Specifically, we first split the concatenated input $\tilde{T}$ into $n$ sentences $\{\tilde{\boldsymbol{t}}_i\}_{i=1}^n$. Each sentence is independently encoded using a pretrained encoder-based language model, and the resulting representations are averaged to form the final representation:
\begin{align}
    \tilde{\mathbf{t}} = \frac{1}{n} \sum_{i=1}^{n} \mathrm{encoder}(\tilde{\boldsymbol{t}}_i),
\end{align}
where $\mathrm{encoder}(\cdot)$ denotes a sentence-level encoder; we use DeBERTa~\citep{DBLP:conf/iclr/HeGC23} in our implementation.

This strategy enables the detector to process inputs of arbitrary length and removes sensitivity to rationale ordering due to the symmetric averaging operator. As a result, \ours{} can robustly incorporate multiple rationales to enhance misinformation detection.

\subsection{CoT-Based Rationale Diversification}
Incorporating foundation model generated news analyses has shown promise in assessing the veracity of news articles~\citep{nan2024let,wu2024fake}. Moreover, using multiple perspectives can provide complementary insights, which may further benefit misinformation detection. To this end, we design five chain-of-thought (CoT) prompts across three categories, \textbf{textual content}, \textbf{visual content}, and \textbf{cross-modal consistency}, to generate a diverse set of veracity-related rationales $\{\bar{R}_j\}_{j=1}^{M}$.

Each rationale $\bar{R}_j$ is generated through a structured multi-turn interaction. First, the LVLM is prompted to analyze the news article from a designated perspective (e.g., ``\emph{analyze the sentiment of this news article}''), yielding an intermediate response $\bar{R}_{j(0)}$. Then, the model is asked to assess the veracity of the article based on this analysis and to justify its judgment, producing $\bar{R}_{j(1)}$. The full rationale $\bar{R}_j$ is formed by concatenating $\bar{R}_{j(0)}$ and $\bar{R}_{j(1)}$.

To encourage reasoning diversity, we design five prompts. For textual content, the prompts guide the LVLM to examine linguistic signals relevant to misinformation, including (\romannumeral 1) \emph{sentiment analysis}~\citep{toughrai2025fake} and (\romannumeral 2) \emph{propaganda tactics}~\citep{piskorski2023semeval}. For visual content, the prompts focus on understanding the accompanying image, specifically through (\romannumeral 3) \emph{object identification}~\citep{ma2024event} and (\romannumeral 4) \emph{image description}~\citep{abdali2024multi}. Lastly, to capture cross-modal consistency, we include (\romannumeral 5) a prompt that evaluates the alignment between textual and visual information, following prior work~\citep{DBLP:conf/iclr/Liu0LHXCHDH25}.
Detailed prompts are provided in the Appendix.

Rather than aiming to exhaustively explore the prompt design space, we intentionally select these five representative prompts to demonstrate the potential of reasoning diversity and the adaptability of \ours{}. The prompts elicit complementary perspectives across textual, visual, and cross-modal dimensions and can be readily extended or customized to suit other reasoning needs or domains.

\begin{table*}[t]
    \centering
    \setlength{\tabcolsep}{1.5mm}
    \small
    \begin{tabular}{cl|cc|cc|cc|cc}
        \toprule[1.5pt]
        \multicolumn{2}{c|}{\multirow{2}{*}{Methods}}&\multicolumn{2}{c|}{Fakeddit}&\multicolumn{2}{c|}{FakeNewsNet}&\multicolumn{2}{c|}{FineFake}&\multicolumn{2}{c}{MMFakeBench}\\
        &&MiF.&MaF.&MiF.&MaF.&MiF.&MaF.&MiF.&MaF.\\
        \midrule[1pt]
        \multirow{8}{*}{Vanilla LVLMs}&InternVL Zero-Shot&$70.1_{\pm 2.4}$&$68.9_{\pm 2.9}$&$74.5_{\pm 3.4}$&$69.9_{\pm 3.3}$&$70.8_{\pm 2.1}$&$70.7_{\pm 2.1}$&$77.5_{\pm 2.5}$&$77.5_{\pm 2.4}$\\
        &InternVL Few-Shot&$70.1_{\pm 3.6}$&$69.6_{\pm 3.8}$&$77.0_{\pm 2.2}$&$72.0_{\pm 3.1}$&$71.4_{\pm 2.7}$&$71.4_{\pm 2.7}$&$72.4_{\pm 2.7}$&$72.2_{\pm 2.6}$\\
        &InternVL Retrieval&$60.6_{\pm 3.3}$&$55.4_{\pm 3.2}$&$64.1_{\pm 2.9}$&$61.1_{\pm 3.3}$&$70.5_{\pm 1.3}$&$70.4_{\pm 1.3}$&$63.6_{\pm 2.9}$&$62.1_{\pm 3.0}$\\
        &InternVL Self-Refine&$59.7_{\pm 3.0}$&$57.9_{\pm 3.0}$&$68.9_{\pm 2.8}$&$65.5_{\pm 2.3}$&$67.3_{\pm 2.8}$&$67.3_{\pm 2.8}$&$64.6_{\pm 1.7}$&$64.6_{\pm 1.7}$\\
        &GPT-4o Zero-Shot&$78.1_{\pm 1.5}$&$78.0_{\pm 1.4}$&$84.0_{\pm 1.8}$&$77.3_{\pm 3.0}$&$75.3_{\pm 3.4}$&$74.5_{\pm 3.6}$&$80.9_{\pm 3.4}$&$80.7_{\pm 3.4}$\\
        &GPT-4o Few-Shot&$78.9_{\pm 2.3}$&$78.8_{\pm 2.3}$&$80.3_{\pm 1.4}$&$72.7_{\pm 3.1}$&$\underline{77.3}_{\pm 0.9}$&$\underline{76.8}_{\pm 0.9}$&$82.3_{\pm 3.1}$&$82.2_{\pm 3.0}$\\
        &GPT-4o Retrieval&$64.1_{\pm 3.6}$&$63.4_{\pm 3.9}$&$81.8_{\pm 1.7}$&$74.8_{\pm 2.3}$&$74.5_{\pm 1.8}$&$73.9_{\pm 1.8}$&$72.9_{\pm 3.4}$&$72.9_{\pm 3.4}$\\
        &GPT-4o Self-Refine&$77.6_{\pm 0.4}$&$77.5_{\pm 0.4}$&$81.2_{\pm 0.6}$&$73.8_{\pm 1.5}$&$73.2_{\pm 3.0}$&$72.2_{\pm 2.9}$&$78.2_{\pm 3.3}$&$78.1_{\pm 3.3}$\\
        \midrule[1pt]
        \multirow{2}{*}{Enhanced LVLMs}&MMD-Agent&$68.9_{\pm 1.9}$&$68.8_{\pm 1.9}$&$67.4_{\pm 4.2}$&$60.4_{\pm 4.5}$&$64.1_{\pm 2.4}$&$64.1_{\pm 2.4}$&$75.3_{\pm 4.2}$&$75.1_{\pm 4.1}$\\
        &Knowledge Card&$52.1_{\pm 3.7}$&$42.4_{\pm 3.6}$&$73.8_{\pm 3.4}$&$67.7_{\pm 5.3}$&$64.5_{\pm 2.2}$&$64.4_{\pm 2.2}$&$57.3_{\pm 2.3}$&$56.3_{\pm 2.7}$\\
        \midrule[1pt]
        \multirow{3}{*}{Trainable Detectors}&CLIP&$86.0_{\pm 2.4}$&$85.9_{\pm 2.4}$&$86.6_{\pm 1.6}$&$82.3_{\pm 1.4}$&$75.7_{\pm 3.3}$&$75.5_{\pm 3.4}$&$84.3_{\pm 2.4}$&$84.2_{\pm 2.4}$\\
        &CAFE&$\underline{87.4}_{\pm 2.1}$&$\underline{87.4}_{\pm 2.1}$&$86.8_{\pm 0.8}$&$82.8_{\pm 1.2}$&$76.2_{\pm 2.7}$&$76.0_{\pm 2.7}$&$\underline{85.4}_{\pm 2.7}$&$\underline{85.4}_{\pm 2.7}$\\
        &COOLANT&$86.4_{\pm 2.3}$&$86.3_{\pm 2.3}$&$85.7_{\pm 1.7}$&$81.3_{\pm 1.9}$&$76.2_{\pm 2.1}$&$76.1_{\pm 2.1}$&$83.2_{\pm 2.1}$&$83.1_{\pm 2.1}$\\
        \midrule[1pt]
        \multirow{2}{*}{LVLM-as-Enhancer}&EARAM&$82.6_{\pm 1.9}$&$82.5_{\pm 1.9}$&$82.9_{\pm 2.9}$&$77.5_{\pm 3.0}$&$73.8_{\pm 2.4}$&$73.7_{\pm 2.3}$&$78.9_{\pm 2.0}$&$78.8_{\pm 2.1}$\\
        &EFND&$80.3_{\pm 1.3}$&$80.2_{\pm 1.2}$&$\underline{87.6}_{\pm 1.0}$&$\underline{84.1}_{\pm 2.0}$&$75.9_{\pm 2.3}$&$75.7_{\pm 2.3}$&$76.5_{\pm 2.5}$&$76.1_{\pm 3.2}$\\
        \midrule[1pt]
        \multicolumn{2}{c|}{\ours{}}&$\textbf{90.8}_{\pm 2.1}$&$\textbf{90.8}_{\pm 2.1}$&$\textbf{89.3}_{\pm 1.9}$&$\textbf{85.5}_{\pm 2.7}$&$\textbf{81.2}_{\pm 1.6}$&$\textbf{81.1}_{\pm 1.7}$&$\textbf{90.4}_{\pm 1.0}$&$\textbf{90.4}_{\pm 1.0}$\\
        \bottomrule[1.5pt]
    \end{tabular}
    \caption{Performance of \ours{} and baselines on four widely used multimodal misinformation detection datasets. ``MiF.'' and ``MaF.'' denote micro- and macro-averaged F1 scores, respectively. \textbf{Bold} indicates the best performance, and \underline{underline} indicates the second-best. \ours{} achieves consistent improvements over state-of-the-art baselines, with gains of up to 5.9\%.}
    \label{tab: main results}
\end{table*}

\subsection{Post-Hoc Rationale Refinement}

While it is possible to directly treat each original rationale $\bar{R}$ as the final input, this approach suffers from two key issues: lack of factuality~\citep{pan2023risk} and lack of relevance~\citep{zheng2025predictions}. To mitigate these challenges, we apply a sentence-level filtering procedure. Specifically, we first split $\bar{R}$ into $\bar{m}$ sentences $\{\bar{\boldsymbol{r}}_k\}_{k=1}^{\bar{m}}$ and apply two filtering strategies to obtain a refined set $\{\boldsymbol{r}_k\}_{k=1}^{m}$ containing only factual and relevant sentences. The final rationale $R$ is obtained by concatenating these filtered sentences.

\paragraph{Factuality Filter.} 
Due to hallucinations~\citep{ji2023survey}, LVLMs may generate sentences with factual errors that degrade downstream detection performance. To address this, we compute a factuality score $s_f(\boldsymbol{r})$ for each sentence $\boldsymbol{r}$ and discard those with low scores. Following~\citet{min2023factscore}, we rely on an external knowledge source (specifically, Wikipedia) to support this evaluation. For each sentence $\boldsymbol{r}$, we retrieve $p$ candidate documents $\{\boldsymbol{d}_i\}_{i=1}^{p}$ and define the factuality score as:
\begin{align}
    s_f(\boldsymbol{r}) = \max_{1 \leq i \leq p} \mathrm{fact}(\boldsymbol{r} \mid \boldsymbol{d}_i),
\end{align}
where $\mathrm{fact}(\boldsymbol{r} \mid \boldsymbol{d}_i)$ quantifies the factual alignment between $\boldsymbol{r}$ and document $\boldsymbol{d}_i$. We compute this score by averaging two complementary signals: a stance classifier~\citep{schuster2021get}, which evaluates how strongly the document supports the sentence, and a summarization precision model~\citep{DBLP:conf/emnlp/FengBBT23}, which assesses how well the sentence summarizes the document:
\begin{align}
    \mathrm{fact}(\boldsymbol{r} \mid \boldsymbol{d}_i) = \frac{1}{2} \left( \mathrm{stance}(\boldsymbol{r}, \boldsymbol{d}_i) + \mathrm{summary}(\boldsymbol{r}, \boldsymbol{d}_i) \right).
\end{align}
Both scores are normalized to the range [0, 1], with higher values indicating stronger factual consistency. We retain the top 50\% of sentences based on the factuality scores: 
\begin{align}
\bar{\boldsymbol{r}}_k \in R \quad \text{if} \quad k \in \mathrm{top}\text{-}50\%_{\ell}(s_f(\bar{\boldsymbol{r}}_\ell)).
\end{align}

\paragraph{Relevance Filter.}
In some cases, the LVLM-generated rationales may diverge from the prompt and include content that is tangential or irrelevant to the source article. To ensure rationale relevance, we assess sentence-level relevance to the input text $T$ by computing semantic similarity. Specifically, we adopt a widely used approach~\citep{lewis2020retrieval} that utilizes an encoder-based language model to obtain sentence embeddings and measure similarity via cosine distance:
\begin{align}
    s_r(T, \boldsymbol{r}) = \cos\left( \mathrm{encoder}(T), \mathrm{encoder}(\boldsymbol{r}) \right),
\end{align}
where $\mathrm{encoder}(\cdot)$ is instantiated as MPNet~\citep{song2020mpnet}. Higher values indicate stronger semantic alignment between the rationale and the input article, where we also retain the top 50\% of sentences.

\section{Experiments}

\subsection{Experimental Setup}
\noindent\textbf{Datasets.} We evaluate \ours{} and existing baselines with four popular datasets: Fakeddit~\citep{DBLP:conf/lrec/NakamuraLW20}, FakeNewsNet~\citep{shu2020fakenewsnet}, FineFake~\citep{zhou2024finefake}, and MMFakeBench~\citep{DBLP:conf/iclr/Liu0LHXCHDH25}. To obtain a robust evaluation, we conduct a five-fold evaluation and report the mean and variance of the performance. Detailed information about datasets is provided in the Appendix. 

\noindent\textbf{Baselines.} We compare \ours{} with four types of state-of-the-art baselines: (\romannumeral 1) Vanilla LVLMs: InternVL V3~\citep{zhu2025internvl3} and GPT-4o with zero-shot, few-shot, retrieval~\citep{lewis2020retrieval}, and self-refine~\citep{madaan2023self} prompt; (\romannumeral 2) Enhanced LVLMs: MMD-Agent~\citep{DBLP:conf/iclr/Liu0LHXCHDH25} and Knowledge Card~\citep{feng2024knowledge}; (\romannumeral 3) Trainable detectors: CLIP~\citep{radford2021learning}, CAFE~\citep{chen2022cross}, and COOLANT~\citep{wang2023cross}; and (\romannumeral 4) LVLM-as-Enhancer: EARAM~\citep{zheng2025predictions} and EFND~\citep{wang2024explainable}. Detailed information about the baselines is provided in the Appendix.

\noindent\textbf{Settings.} We use GPT-4o as the primary LVLM backbone for \ours{}. To ensure fair comparison, all detectors are evaluated under consistent hyperparameter settings across folds. For LVLM inference, we disable sampling by either setting the temperature to zero or configuring \texttt{do\_sample} to \texttt{False}, ensuring deterministic outputs and reproducibility. Additional implementation details and experimental configurations are provided in the Appendix.

\begin{table*}[t]
    \centering
    \setlength{\tabcolsep}{1.5mm}
    \small
    \begin{tabular}{cl|c|c|c|c}
        \toprule[1.5pt]
        Models&Variants&Fakeddit&FakeNewsNet&FineFake&MMFakeBench\\
        \midrule[1pt]
        \multirow{6}{*}{CLIP}&Original&$86.0_{\pm 2.4}$&$86.6_{\pm 1.6}$&$75.7_{\pm 3.3}$&$84.3_{\pm 2.4}$\\
        &MMD-Agent&$82.5_{\pm 2.9}\ (4.1\%\downarrow)$&$83.4_{\pm 1.7}\ (3.8\%\downarrow)$&$71.7_{\pm 2.7}\ (5.3\%\downarrow)$&$82.2_{\pm 1.4}\ (2.5\%\downarrow)$\\
        &Knowledge Card&$84.1_{\pm 3.6}\ (2.2\%\downarrow)$&$84.5_{\pm 2.6}\ (2.5\%\downarrow)$&$74.6_{\pm 2.1}\ (1.5\%\downarrow)$&$84.2_{\pm 2.5}\ (0.1\%\downarrow)$\\
        &EARAM&$82.6_{\pm 2.5}\ (4.0\%\downarrow)$&$83.8_{\pm 2.2}\ (3.3\%\downarrow)$&$73.4_{\pm 2.3}\ (3.0\%\downarrow)$&$81.8_{\pm 1.6}\ (3.0\%\downarrow)$\\
        &EFND&$83.8_{\pm 2.8}\ (2.6\%\downarrow)$&$84.4_{\pm 1.9}\ (2.6\%\downarrow)$&$73.4_{\pm 1.7}\ (3.0\%\downarrow)$&$82.7_{\pm 1.8}\ (1.9\%\downarrow)$\\
        &\ours{}&\cellcolor{cellgray}$85.3_{\pm 2.2}\ (0.8\%\downarrow)$&\cellcolor{cellgray}$84.6_{\pm 1.9}\ (2.3\%\downarrow)$&\cellcolor{cellgray}$77.1_{\pm 2.1}\ (1.8\%\uparrow)$&\cellcolor{cellgray}$85.2_{\pm 1.7}\ (1.1\%\uparrow)$\\
        \midrule[1pt]
        \multirow{6}{*}{CAFE}&Original&$87.4_{\pm 2.1}$&$86.8_{\pm 0.8}$&$76.2_{\pm 2.7}$&$85.4_{\pm 2.7}$\\
        &MMD-Agent&$85.8_{\pm 2.2}\ (1.8\%\downarrow)$&$88.4_{\pm 1.1}\ (1.9\%\downarrow)$&$74.1_{\pm 2.0}\ (2.8\%\downarrow)$&$84.6_{\pm 2.4}\ (0.9\%\downarrow)$\\
        &Knowledge Card&$85.4_{\pm 3.0}\ (2.3\%\downarrow)$&$88.5_{\pm 2.1}\ (2.0\%\uparrow)$&$76.6_{\pm 1.8}\ (0.5\%\uparrow)$&$85.3_{\pm 2.7}\ (0.1\%\downarrow)$\\
        &EARAM&$85.8_{\pm 1.3}\ (1.8\%\downarrow)$&$87.6_{\pm 0.9}\ (0.9    \%\uparrow)$&$73.9_{\pm 2.3}\ (3.0\%\downarrow)$&$84.7_{\pm 2.5}\ (0.8\%\downarrow)$\\
        &EFND&$86.5_{\pm 1.6}\ (1.0\%\downarrow)$&$88.5_{\pm 0.9}\ (2.0    \%\downarrow)$&$75.0_{\pm 1.8}\ (1.6\%\downarrow)$&$84.9_{\pm 2.9}\ (0.6\%\downarrow)$\\
        &\ours{}&\cellcolor{cellgray}$90.5_{\pm 2.0}\ (3.5\%\uparrow)$&\cellcolor{cellgray}$88.8_{\pm 1.6}\ (2.3\%\uparrow)$&\cellcolor{cellgray}$80.2_{\pm 1.9}\ (5.2\%\uparrow)$&\cellcolor{cellgray}$88.6_{\pm 1.6}\ (3.7\%\uparrow)$\\
        \midrule[1pt]
        \multirow{6}{*}{COOLANT}&Original&$86.4_{\pm 2.3}$&$85.7_{\pm 1.7}$&$76.2_{\pm 2.1}$&$83.2_{\pm 2.1}$\\
        &MMD-Agent&$85.5_{\pm 2.5}\ (1.0\%\downarrow)$&$87.5_{\pm 1.1}\ (2.1\%\uparrow)$&$75.2_{\pm 2.3}\ (1.3\%\downarrow)$&$85.3_{\pm 2.3}\ (2.5\%\uparrow)$\\
        &Knowledge Card&$85.6_{\pm 2.8}\ (0.9\%\downarrow)$&$89.2_{\pm 1.7}\ (4.1\%\uparrow)$&$78.2_{\pm 2.4}\ (2.6\%\uparrow)$&$85.0_{\pm 2.9}\ (2.2\%\uparrow)$\\
        &EARAM&$84.5_{\pm 2.4}\ (2.2\%\downarrow)$&$87.9_{\pm 1.0}\ (2.6\%\uparrow)$&$76.0_{\pm 2.7}\ (0.3\%\downarrow)$&$85.8_{\pm 1.5}\ (3.1\%\uparrow)$\\
        &EFND&$86.9_{\pm 1.6}\ (0.6\%\uparrow)$&$88.7_{\pm 1.3}\ (3.5\%\uparrow)$&$76.4_{\pm 2.3}\ (0.3\%\uparrow)$&$85.1_{\pm 2.6}\ (2.3\%\uparrow)$\\
        &\ours{}&\cellcolor{cellgray}$90.8_{\pm 2.1}\ (5.1\%\uparrow)$&\cellcolor{cellgray}$89.3_{\pm 1.9}\ (4.3\%\uparrow)$&\cellcolor{cellgray}$81.2_{\pm 1.6}\ (6.6\%\uparrow)$&\cellcolor{cellgray}$90.4_{\pm 1.0}\ (8.7\%\uparrow)$\\
        \bottomrule[1.5pt]
    \end{tabular}
    \caption{Micro-averaged F1 scores of trainable detectors enhanced with \ours{} and baseline methods. The best performance in each setting is highlighted. \ours{} improves detector performance by up to 8.7\%, demonstrating its effectiveness in addressing the limitations of rationale diversity, factuality, and relevance.}
    \label{tab: PnP}
\end{table*}

\subsection{Effectiveness of \ours{}}
We present the performance of \ours{} and the state-of-the-art baseline in Table~\ref{tab: main results}.

\noindent\textbf{Trainable detectors remain highly competitive.} Among all baselines, supervised trainable detectors exhibit the strongest standalone performance, coming close to \ours{} across benchmarks. This highlights the effectiveness of learned representations under direct supervision and further supports the value of integrating LVLM-generated rationales into such architectures.

\noindent\textbf{Existing LVLM-as-Enhancer approaches face generalization issues.} Interestingly, existing LVLM-as-Enhancer methods often underperform, sometimes even falling below vanilla LVLMs. This suggests poor generalization across datasets, likely due to narrow reasoning scopes and low-quality rationales. For instance, EARAM focuses only on commonsense and complementarity, while EFND centers on debate-style veracity analysis. Both approaches lack broad perspective integration. We also observe that non-factual or off-topic content in generated rationales may mislead the detector. These findings directly motivate our design of \ours{}, which addresses these gaps by enhancing the diversity, factuality, and relevance of LVLM-generated rationales.

\noindent\textbf{\ours{} achieves state-of-the-art-performance.} \ours{} consistently outperforms the strongest baseline on all four datasets, achieving gains of 1.9\% to 5.9\% in micro-averaged F1 score. Notably, both vanilla and enhanced LVLMs underperform on most benchmarks, suggesting that LVLMs alone struggle with factual reasoning and precise veracity assessment. These results reinforce the importance of the \emph{LVLM-as-Enhancer} paradigm, where external reasoning is paired with trainable detectors.

\begin{figure}[t]
    \centering
    \includegraphics[width=0.8\linewidth]{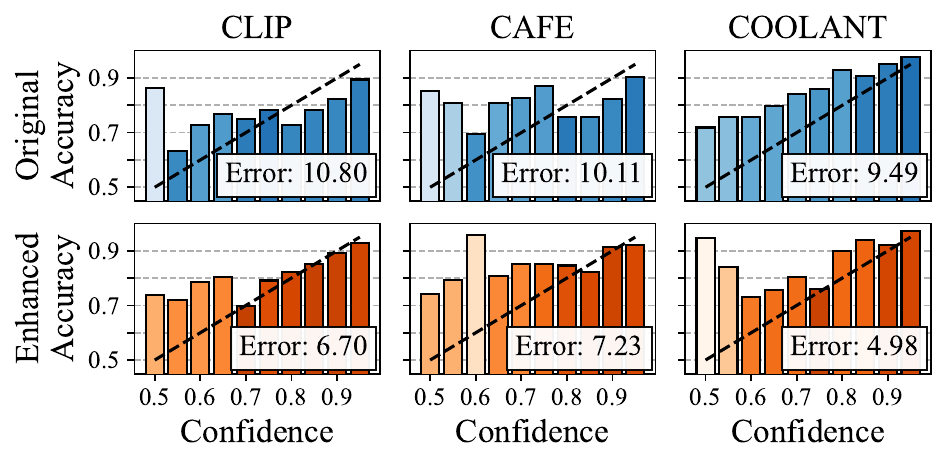}
    \caption{Calibration performance of existing trainable detectors with and without \ours{} enhancement. ``Error'' refers to the expected calibration error (ECE, $\times 100$), where lower values indicate better calibration. \ours{} not only improves detection accuracy but also enhances the reliability of confidence estimates.}
    \label{fig: calibration}
\end{figure}

\subsection{Adaptability to Diverse Detectors}
\ours{} is designed to be compatible with any trainable detector and any set of LVLM-generated rationales. To evaluate its adaptability, we assess the effectiveness of \ours{}-generated rationales when integrated with various detectors. 

As shown in Table~\ref{tab: PnP}, we draw three key observations. (\romannumeral 1) \ours{} significantly improves the performance of all trainable detectors, achieving gains of up to 8.7\%. It demonstrates its effectiveness in addressing key limitations of the LVLM-as-Enhancer paradigm and enhancing detection performance across diverse architectures. (\romannumeral 2) Performance gains gradually decrease across COOLANT, CAFE, and CLIP, with a slight performance drop on CLIP after enhancement. It suggests that more complex detector architectures may be more effective at extracting signal from rationales. It trend aligns with prior work emphasizing the role of architectural design in driving performance. (\romannumeral 3) Rationales generated by baseline methods often fail to improve detector performance and, in some cases, lead to degradation. It highlights the importance of rationale quality, without sufficient diversity, factuality, and relevance, even strong detectors cannot reliably benefit from external explanations.

Beyond accuracy, we also evaluate the \emph{credibility} of the detectors before and after enhancement using \ours{}. We quantify this using expected calibration error (ECE)~\citep{guo2017calibration}. As shown in Figure~\ref{fig: calibration}, detectors become better calibrated after incorporating \ours{}, with reductions in ECE of up to 47.5\%. These results indicate that \ours{} not only improves prediction performance but also enhances the reliability of model confidence estimates.

\subsection{Ablation Study}

We conduct an ablation study to evaluate the contribution of each component in \ours{}. Specifically, we consider four settings: (\romannumeral 1) replacing the five diverse CoT prompts with a single fixed CoT prompt; (\romannumeral 2) removing both the factuality and relevance filtering strategies; (\romannumeral 3) prompting the LVLM with a generic instruction to generate rationales (i.e., ``\textit{Determine whether this news with text and image is fake or real. Meanwhile, provide a comprehensive explanation.}''); and (\romannumeral 4) replacing GPT-4o with the open-sourced InternVL V3 model. We report results on the FineFake dataset in Table~\ref{tab: ablation}, with full results provided in the Appendix. The findings show that detectors fail to benefit from rationales generated by a vanilla prompt, underscoring the importance of rationale diversity, factuality, and relevance. Moreover, ablating any individual component of \ours{} leads to a performance drop in most cases, with decreases of up to 5.8\%. These results confirm that each module in \ours{} plays an essential role in overcoming the limitations of the LVLM-as-Enhancer paradigm.

\begin{table}[t]
    \centering
    \setlength{\tabcolsep}{1.5mm}
    \small
    \begin{tabular}{l|c|c|c}
        \toprule[1.5pt]
        Variants&CLIP&CAFE&COOLANT\\
        \midrule[1pt]
        \ours{}&$77.1_{\pm 2.1}$&$80.2_{\pm 1.9}$&$81.2_{\pm 1.6}$\\
        \midrule[1pt]
        \multirow{2}{*}{w/o Multiple}&$72.6_{\pm 3.8}$&$78.3_{\pm 1.9}$&$79.6_{\pm 1.4}$\\
        &$5.8\%\downarrow$&$2.4\%\downarrow$&$2.0\%\downarrow$\\
        \multirow{2}{*}{w/o Filter}&$72.7_{\pm 3.5}$&$78.0_{\pm 2.3}$&$78.9_{\pm 1.4}$\\
        &$5.7\%\downarrow$&$2.7\%\downarrow$&$2.8\%\downarrow$\\
        \multirow{2}{*}{w/ Vanilla}&$73.0_{\pm 2.1}$&$73.8_{\pm 2.3}$&$76.7_{\pm 2.4}$\\
        &$5.3\%\downarrow$&$8.0\%\downarrow$&$5.5\%\downarrow$\\
        \multirow{2}{*}{w/ InternVL}&$77.5_{\pm 2.2}$&$78.8_{\pm 1.3}$&$79.9_{\pm 2.3}$\\
        &$0.5\%\uparrow$&$1.7\%\downarrow$&$1.6\%\downarrow$\\
        \bottomrule[1.5pt]
    \end{tabular}
    \caption{Ablation study of \ours{}. ``w/o Multiple'' uses a single specific CoT prompt instead of five; ``w/o Filter'' removes the factuality and relevance filters; ``w/ Vanilla'' uses rationales generated from a simple prompt; and ``w/ InternVL'' replaces GPT-4o with InternVL V3. Results show that each component contributes to performance gains.}
    \label{tab: ablation}
\end{table}

\section{Rationale Quality Analysis}
We further evaluate whether \ours{} effectively addresses the challenges of limited diversity, factuality, and relevance in generated rationales.

\subsection{Overall Helpfulness Evaluation}

We begin by assessing the overall quality of rationales generated by \ours{}. To this end, we conduct a human evaluation with three experts in misinformation-related topics. Four types of rationales are selected for comparison: (\romannumeral 1) \textbf{\ours{}}: rationales generated by \ours{} before filtering; (\romannumeral 2) \textbf{Baseline}: rationales generated by EFND; (\romannumeral 3) \textbf{Single}: rationales generated from a single randomly selected perspective; and (\romannumeral 4) \textbf{Filtered}: rationales produced by \ours{} after applying factuality and relevance filters.

We conduct four pairwise comparisons: \ours{} vs. Baseline, \ours{} vs. Single, Filtered vs. Baseline, and Filtered vs. Single. For each pair, experts are asked to judge which rationale is more helpful for verifying the veracity of the news article, or to indicate if the two are indistinguishable. Final decisions are determined via majority vote. Details of the evaluation protocol are provided in the Appendix. Fleiss’ Kappa across all judgments is 0.34, indicating a fair level of inter-rater agreement.

\begin{figure}[t]
    \centering
    \includegraphics[width=0.8\linewidth]{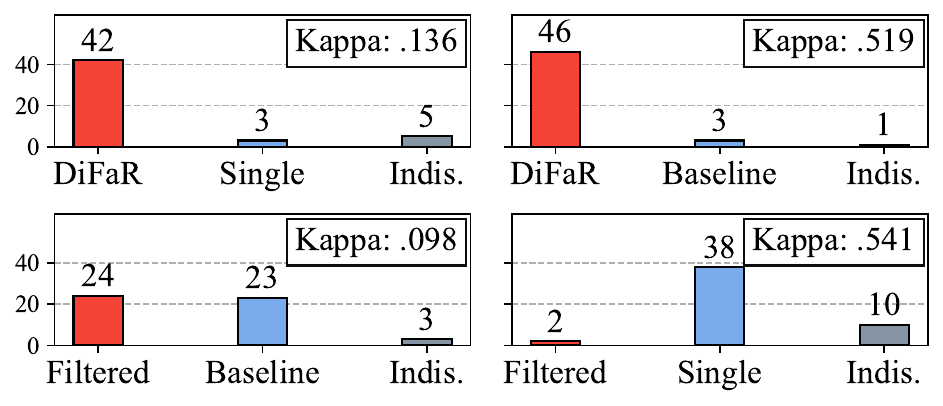}
    \caption{Human voting results of generated rationales via pairwise comparison. ``Kappa'' denotes the Fleiss’ Kappa score among three experts, and ``Indis.'' indicates the proportion of instances where the two rationales were judged indistinguishable. \ours{} produces the most helpful rationales for veracity assessment.}
    \label{fig: human_evaluation}
\end{figure}

Results in Figure~\ref{fig: human_evaluation} show that \ours{} significantly outperforms both the Baseline and Single settings, demonstrating that it produces more useful rationales for human misinformation assessment. However, after post-hoc filtering, the advantage of \ours{} is reduced and, in some cases, performs worse than the Single baseline. We speculate that while filtering improves factuality and relevance, it may also degrade the fluency and coherence of the rationales, limiting their interpretability for human readers. 

\subsection{Fine-Grained Quality Evaluation}

\paragraph{Diversity.} 
\ours{} employs five chain-of-thought (CoT) prompts to capture signals from diverse reasoning perspectives. As a baseline, we compare against EFND, which generates rationales from only two perspectives. We begin by computing the proportion of distinct tokens in rationales generated on the FineFake dataset. \ours{} achieves a distinct token ratio of 0.904, substantially higher than EFND's 0.406, suggesting broader lexical coverage.

To further quantify diversity, we analyze token frequency and inter-prompt similarity. Specifically, we use infini-gram~\citep{liu2024infini} to measure token frequency and BERTScore~\citep{DBLP:conf/iclr/ZhangKWWA20} to compute pairwise similarity between rationales generated from different prompts. Lower values on both metrics indicate greater diversity. As shown in the Appendix, \ours{} achieves a lower average similarity (0.57 vs. 0.66) and lower token frequency ($2.41 \times 10^6$ vs. $2.85 \times 10^6$), confirming that it produces more lexically diverse outputs.

\begin{table}[t]
    \centering
    \setlength{\tabcolsep}{1.5mm}
    \small
    \begin{tabular}{c|c|c|c|c}
        \toprule[1.5pt]
        Dataset&Orig.&Consis.&Textual&Visual\\
        \midrule[1pt]
        \multirow{2}{*}{Fakeddit}&\multirow{2}{*}{$90.8_{\pm2.1}$}&$89.5_{\pm1.7}$&$88.8_{\pm3.1}$&$90.2_{\pm1.5}$\\
        &&$1.4\%\downarrow$&$2.2\%\downarrow$&$0.7\%\downarrow$\\
        \midrule[1pt]
        \multirow{2}{*}{FakeNewsNet}&\multirow{2}{*}{$89.3_{\pm1.9}$}&$88.8_{\pm1.6}$&$88.9_{\pm2.2}$&$89.3_{\pm2.0}$\\
        &&$0.6\%\downarrow$&$0.5\%\downarrow$&$0.0\%\downarrow$\\
        \midrule[1pt]
        \multirow{2}{*}{FineFake}&\multirow{2}{*}{$81.2_{\pm1.6}$}&$80.0_{\pm1.7}$&$79.4_{\pm1.7}$&$80.8_{\pm1.5}$\\
        &&$1.5\%\downarrow$&$2.2\%\downarrow$&$0.5\%\downarrow$\\
        \midrule[1pt]
        \multirow{2}{*}{MMFakeBench}&\multirow{2}{*}{$90.4_{\pm1.0}$}&$91.0_{\pm1.4}$&$88.1_{\pm2.8}$&$89.6_{\pm2.0}$\\
        &&$0.7\%\uparrow$&$2.5\%\downarrow$&$0.9\%\downarrow$\\
        \bottomrule[1.5pt]
        
    \end{tabular}
    \caption{Ablation study on CoT prompt categories, where only rationales from a single prompt type are retained. ``Orig.'' denotes the original performance of \ours{} using all five prompts, and ``Consis.'' refers to the variant using only the cross-modal consistency prompt. We report micro-averaged F1 scores and the corresponding performance changes. Results show that diverse rationales generally outperform those from any single perspective.}
    \label{tab: fine_grained}
\end{table}

We also conduct an ablation study with COOLANT to evaluate the impact of different CoT prompt types. Table~\ref{tab: fine_grained} shows that removing any single prompt category leads to performance drop of up to 2.5\%, validating the importance of incorporating multiple reasoning perspectives for misinformation detection. These findings support our design of rationale diversification as a core component of \ours{}.

\begin{figure}[t]
    \centering
    \includegraphics[width=0.8\linewidth]{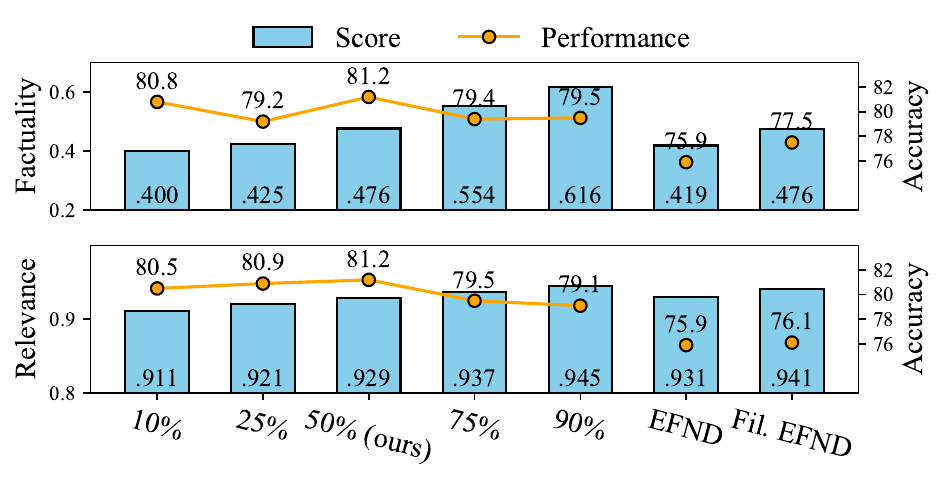}
    \caption{Performance of \ours{} under different filtering thresholds, along with corresponding factuality and relevance scores. ``Fil. EFND'' denotes the EFND baseline after filtering out the bottom 25\% of sentences based on scores. It shows that moderate filtering improves performance, while overly aggressive filtering may reduce effectiveness.}
    \label{fig: scores}
\end{figure}

\paragraph{Factuality and Relevance.} 
\ours{} employs two post-hoc filtering strategies to improve the factuality and relevance of generated rationales. To investigate their effect, we vary the filtering threshold and evaluate both model performance and average factuality/relevance scores, using EFND as a baseline for comparison. Figure~\ref{fig: scores} shows that increasing the filtering threshold results in higher factuality and relevance scores. However, the overall detection performance of \ours{} does not continue to increase proportionally. We speculate that overly aggressive filtering, while improving quality scores, may remove semantically rich content, thereby weakening the enhancement signal provided to detectors.

Interestingly, we observe that EFND and \ours{} with a low filtering threshold (e.g., 10\%) achieve similar factuality and relevance scores, yet their detection performance differs significantly. We attribute this to the presence of low-quality sentences in EFND’s rationales, which likely mislead the detector. To test this, we apply our filtering strategy to EFND and remove the bottom 25\% of its sentences by factuality and relevance score. This leads to a measurable performance gain, supporting the effectiveness of our filtering design. Moreover, unlike EFND, \ours{} generates rationales from diverse prompts, offering broader perspectives and richer semantic coverage. This diversity allows \ours{} to retain informative content even after filtering, contributing to its superior performance.

\subsection{Case Study}
\begin{figure}[t]
    \centering
    \includegraphics[width=0.8\linewidth]{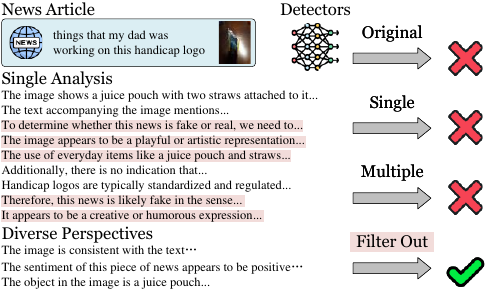}
    \caption{Case study of predictions using different variants of \ours{} with COOLANT as the base detector. The original detector fails to identify the misinformation, while enhancement with \ours{} enables a correct prediction.}
    \label{fig: case_study}
\end{figure}

We analyze a representative example from the dataset to illustrate how \ours{} contributes to improved misinformation detection. The generated rationales and corresponding model predictions are shown in Figure~\ref{fig: case_study}. In this case, the original COOLANT model produces an incorrect prediction. Although a single rationale correctly identifies the misinformation, incorporating it into COOLANT does not lead to a correct prediction. In contrast, when COOLANT is enhanced with the full set of diverse rationales and non-factual or irrelevant sentences are filtered out, the model successfully detects the misinformation. This example demonstrates that \ours{} provides richer and more focused semantic signals, which meaningfully support downstream detection.

\section{Related Work}
Multimodal misinformation detectors typically encode textual content and visual content using pretrained encoders, followed by architectures designed to model cross-modal interactions~\citep{tonglet2024image, tong2024mmdfnd, zhang2024natural, lu2025dammfnd, cao2025external, li2025learning, yu2025racmc, feng2025contradicted}.  
With the rise of LVLMs, early work directly employs LVLMs as backbones to identify misinformation~\citep{lucas2023fighting, gabriel2024misinfoeval, huang2024creation, liu2025fmdllama, chen2025retrieverguard, li2025imrrf,wu2025seeing}. 
However, these models often suffer from hallucinations and lack factual grounding~\citep{hu2024bad}, limiting their effectiveness.
Thus, \emph{LVLM-as-Enhancer} paradigm is proposed. This paradigm first design prompts to generate external textual content, namely, explanations or rationales, to provide rich semantic information~\citep{saha2024integrating, liu2024fka}, such as stance~\citep{choi2025limited}, propagation~\citep{liu2024skepticism}, and entities~\citep{ma2024fake}. They then design a trainable module to capture the semantic information to enhance performance~\citep{zhang2025llms, zhou2025collaborative}. In this work, we identify key limitations of existing LVLM-generated rationales, specifically, their lack of diversity, factuality, and relevance, and propose \ours{}, a general framework that addresses these challenges through multi-perspective prompting and post-hoc filtering.
\section{Conclusion}

We propose \ours{}, a simple yet effective framework under the \emph{LVLM-as-Enhancer} paradigm that seamlessly adapts to multiple rationales without requiring structural changes to the detector.  It employs five chain-of-thought prompts to encourage diverse reasoning and two post-hoc filtering strategies to ensure factuality and relevance. Extensive experiments show that \ours{} achieves state-of-the-art performance and could significantly enhance existing trainable detectors. Further analyses, including human evaluations, confirm that \ours{} successfully enhances rationale diversity, factuality, and relevance.

\bibliography{aaai2026}

\appendix
\section{Prompts of \ours{}}
For the diverse perspectives of news articles, we employ the following prompts (assuming that we first provide the textual and visual content to LVLMs):
\begin{itemize}
    \item \textbf{Sentiment.} \textit{Please analyze the sentiment of this piece of news.}
    \item \textbf{Propaganda.} \textit{Please analyze the propaganda tactics utilized in this piece of news.}
    \item \textbf{Consistency.} \textit{Please analyze the consistency between the text and the image of this piece of news.}
    \item \textbf{Object.} \textit{Please analyze the object that appears in the image of this piece of news.}
    \item \textbf{Description.} \textit{Please describe this image in this piece of news.}
\end{itemize}
Finally, we prompt LVLMs to judge the veracity by prompt \textit{Based on the analysis, determine whether this news with text and image is fake or real. Meanwhile, provide a comprehensive explanation.}

\section{Datasets}
We evaluate \ours{} and existing baselines with four widely used multimodal misinformation detection datasets, where each news article contains text content and an image, including human-written and machine-generated multimodal news articles:
\begin{itemize}
    \item Fakeddit~\citep{DBLP:conf/lrec/NakamuraLW20} is a multimodal dataset consisting of over 1 million samples from multiple categories of fake news, where the source is Reddit. Each instance is labeled according to 2-way, 3-way, and 6-way classification categories through distant supervision. We employ 2-way labels, namely, consider it a binary classification task. 
    \item FakeNewsNet~\citep{shu2020fakenewsnet} contains two comprehensive data sets: Politifact, which includes political news, and Gossipcop, which includes entertainment news. Each instance contains diverse features such as news content, social context, and spatiotemporal information. We only employ the textual content and visual content in news articles. 
    \item FineFake~\citep{zhou2024finefake} includes 16,909 news articles covering six semantic topics and eight platforms. Each instance contains multi-modal content, potential social context, semi-manually verified common knowledge, and fine-grained annotations that surpass conventional binary labels. We only employ the textual content and visual content in news articles, and leverage the binary labels.
    \item MMFakeBench~\citep{DBLP:conf/iclr/Liu0LHXCHDH25} includes three critical sources: textual veracity distortion, visual veracity distortion, and cross-modal consistency distortion, along with 12 sub-categories of misinformation forgery types. It covers machine-generated news articles, including the generated textual content and visual content. We leverage the binary labels. 
\end{itemize}
\begin{table}[t]
    \centering
    \begin{tabular}{c|c|c|c}
         \toprule[1.5pt]
         Datasets&\# Instances&\# Fake&\# Real\\
         \midrule[1pt]
         Fakeddit&1,000&500&500\\
         FakeNewsNet&985&275&710\\
         FineFake&1,000&500&500\\
         MMFakeBench&1,000&500&500\\
         \bottomrule[1.5pt]
    \end{tabular}
    \caption{The statistics of datasets.}
    \label{tab: dataset statistics}
\end{table}

To ensure a fair comparison, we sample from the original datasets to create balanced subsets with a similar number of real and fake instances. The statistics of the datasets are shown in Table~\ref{tab: dataset statistics}. Additionally, to enhance the robustness of the results, we randomly partition each dataset into five equal folds for cross-validation.

\begin{table*}[t]
    \centering
    \setlength{\tabcolsep}{1.5mm}
    \small
    \begin{tabular}{cl|c|c|c|c}
        \toprule[1.5pt]
        Models&Variants&Fakeddit&FakeNewsNet&FineFake&MMFakeBench\\
        \midrule[1pt]
        \multirow{5}{*}{CLIP}&\ours{}&$85.3_{\pm 2.2}$&$84.6_{\pm 1.9}$&$77.1_{\pm 2.1}$&$85.2_{\pm 1.7}$\\
        &w/o Multiple&$84.8_{\pm 2.2}\ (0.6\%\downarrow)$&$83.6_{\pm 2.4}\ (1.2\%\downarrow)$&$72.6_{\pm 3.8}\ (5.8\%\downarrow)$&$81.8_{\pm 1.6}\ (4.0\%\downarrow)$\\
        &w/o Filter&$82.8_{\pm 3.0}\ (2.9\%\downarrow)$&$83.7_{\pm 2.6}\ (1.1\%\downarrow)$&$72.7_{\pm 3.5}\ (5.7\%\downarrow)$&$82.0_{\pm 1.9}\ (3.8\%\downarrow)$\\
        &w/ Vanilla&$83.1_{\pm 3.2}\ (2.6\%\downarrow)$&$83.7_{\pm 2.9}\ (1.1\%\downarrow)$&$73.0_{\pm 2.1}\ (5.3\%\downarrow)$&$81.7_{\pm 2.0}\ (4.1\%\downarrow)$\\
        &w/ InternVL&$83.7_{\pm 3.5}\ (1.9\%\downarrow)$&$84.8_{\pm 1.9}\ (0.2\%\uparrow)$&$77.5_{\pm 2.2}\ (0.5\%\uparrow)$&$85.2_{\pm 2.0}\ (0.0\%\downarrow)$\\
        \midrule[1pt]
        \multirow{5}{*}{CAFE}&\ours{}&$90.5_{\pm 2.0}$&$88.8_{\pm 1.6}$&$80.2_{\pm 1.9}$&$88.6_{\pm 1.6}$\\
        &w/o Multiple&$88.9_{\pm 2.3}\ (1.8\%\downarrow)$&$88.5_{\pm 1.7}\ (0.3\%\downarrow)$&$78.3_{\pm 1.9}\ (2.4\%\downarrow)$&$87.5_{\pm 2.9}\ (1.2\%\downarrow)$\\
        &w/o Filter&$90.8_{\pm 2.7}\ (0.3\%\uparrow)$&$88.9_{\pm 2.2}\ (0.1\%\uparrow)$&$78.0_{\pm 2.3}\ (2.7\%\downarrow)$&$87.6_{\pm 1.6}\ (1.1\%\downarrow)$\\
        &w/ Vanilla&$83.6_{\pm 2.6}\ (7.6\%\downarrow)$&$88.0_{\pm 1.5}\ (0.9\%\downarrow)$&$73.8_{\pm 2.3}\ (8.0\%\downarrow)$&$83.5_{\pm 1.9}\ (5.8\%\downarrow)$\\
        &w/ InternVL&$85.7_{\pm 2.2}\ (5.3\%\downarrow)$&$88.7_{\pm 1.0}\ (0.1\%\downarrow)$&$78.8_{\pm 1.3}\ (1.7\%\downarrow)$&$86.4_{\pm 3.1}\ (2.5\%\downarrow)$\\
        \midrule[1pt]
        \multirow{5}{*}{COOLANT}&\ours{}&$90.8_{\pm 2.1}$&$89.3_{\pm 1.9}$&$81.2_{\pm 1.6}$&$90.4_{\pm 1.0}$\\
        &w/o Multiple&$87.8_{\pm 1.2}\ (3.3\%\downarrow)$&$88.7_{\pm 1.3}\ (0.7\%\downarrow)$&$79.6_{\pm 1.4}\ (2.0\%\downarrow)$&$87.9_{\pm 3.3}\ (2.8\%\downarrow)$\\
        &w/o Filter&$89.9_{\pm 2.5}\ (1.0\%\downarrow)$&$89.2_{\pm 2.0}\ (0.1\%\downarrow)$&$78.9_{\pm 1.4}\ (2.8\%\downarrow)$&$88.5_{\pm 2.2}\ (2.1\%\downarrow)$\\
        &w/ Vanilla&$83.4_{\pm 3.0}\ (8.1\%\downarrow)$&$87.1_{\pm 1.6}\ (2.5\%\downarrow)$&$76.7_{\pm 2.4}\ (5.5\%\downarrow)$&$84.2_{\pm 3.1}\ (6.9\%\downarrow)$\\
        &w/ InternVL&$85.7_{\pm 1.9}\ (5.6\%\downarrow)$&$87.6_{\pm 1.5}\ (1.9\%\downarrow)$&$79.9_{\pm 2.3}\ (1.6\%\downarrow)$&$85.3_{\pm 3.0}\ (5.6\%\downarrow)$\\
        \bottomrule[1.5pt]
    \end{tabular}
    \caption{The ablation study of \ours{}. It illustrates that each module of \ours{} could improve the detection performance.}
    \label{tab: ablation}
\end{table*}

\begin{table}[t]
    \centering
    \small
    \begin{tabular}{c|c|c|c}
        \toprule[1.5pt]
        Hyperparameter&CLIP&CAFE&COOLANT\\
        \midrule[1pt]
        Optimizer&Adam&Adam&AdamW\\
        Weight Decay&1e-5&1e-5&5e-4\\
        Dropout&0.5&0.5&0.5\\
        Learning Rate&1e-3&1e-3&1e-4\\
        Batch Size&256&32&64\\
        \midrule[1pt]
        \bottomrule[1.5pt]
    \end{tabular}
    \caption{The hyperparameters of the baselines. }
    \label{tab: hyperparameters}
\end{table}

\section{Baselines}
We compare \ours{} with four types of state-of-the-art baselines.
\paragraph{Vanilla LVLMs} employs simple prompts to prompt GPT-4o and InternVL V3 to conduct misinformation detection. We employ the following prompt style:
\begin{itemize}
    \item Zero-shot directly prompts LVLMs to obtain the results, where the prompt is as follows:
    
    \noindent
    \textit{Text: {Text}\\Image: {Image}\\Based on the above text and image, please judge whether this piece of news is real or fake. Just output real or fake without any explanations.}
    \item Few-shot additional provides a random sample of instances with the label to LVLMs, where the prompt is as follows:

    \noindent
    \textit{Text: {Text}\\Image: {Image}\\Label: {Label}\\Based on the above examples, please judge whether the following piece of news is real or fake. Just output real or fake without any explanations.\\Text: {Text}\\Image: {Image}\\Label:}
    \item Retrieval~\citep{lewis2020retrieval} first retrieves the three most related news from BBC news resource~\citep{li2024latesteval} using bm25. It then provided the related news to LVLMs as external content, where the prompt is as follows:

    \noindent
    \textit{Related News:\\{Retrieval news articles}\\Text: {Text}\\Image: {Image}\\Based on the above text and image, please judge whether this piece of news is real or fake. Just output real or fake without any explanations.}
    \item Self-Refine~\citep{madaan2023self} prompts LVLMs to check whether the answer is correct and refine the predictions, where the prompt is as follows:
    
    \noindent
    \textit{Text: {Text}\\Image: {Image}\\Based on the above text and image, please judge whether this piece of news is real or fake. Just output real or fake without any explanations.\\{The first prediction}\\Is the answer correct? If not, give your answer.}
    
\end{itemize}

\paragraph{Enhanced LVLMs} design a framework containing multiple prompts to enhance the ability of LVLMs, including:
\begin{itemize}
    \item MMD-Agent~\citep{DBLP:conf/iclr/Liu0LHXCHDH25} could integrate the reasoning, action, and tool-use capabilities of LVLM agents to enhance the generalization and improve the detection performance. 
    \item Knowledge Card~\citep{feng2024knowledge} proposes a modular framework to plug in new factual and relevant knowledge into large language models. We employ the bottom-up approach and the advised cards to enhance LVLMs.
\end{itemize}
\paragraph{Trainable detectors} represent the traditional multimodal misinformation detectors that require to learn the parameters, including:
\begin{itemize}
    \item CLIP~\citep{radford2021learning} is a widely used backbone to encode textual and visual information. We employ MLP layers to classify misinformation after obtaining the representations. 
    \item CAFE~\citep{chen2022cross} is an ambiguity-aware multimodal fake news detection method, containing a cross-modal alignment module, a cross-modal ambiguity learning module, and a cross-modal fusion module.
    \item COOLANT~\citep{wang2023cross} is a cross-modal contrastive learning framework for multimodal fake news detection, including an auxiliary task, a cross-modal fusion module, and an attention mechanism with an attention guidance module.
\end{itemize}
\paragraph{LVLM-as-enhancer} detectors contain two representative baselines with this paradigm, including:
\begin{itemize}
    \item EARAM~\citep{zheng2025predictions} could use multimodal small language models to extract useful rationales from the multi-perspective analyses of LVLMs. The LVLMs analyze the common sense and the complementarity of news articles.
    \item EFND~\citep{wang2024explainable} designs a prompt-based module that utilizes a large language model to generate justifications by inferring reasons towards two possible veracities. It also proposes a specific trainable module to capture the signals from two perspectives. 
\end{itemize}
We believe these four categories of multimodal misinformation detectors encompass most existing approaches and represent the advances in this field.

\section{Settings}
\paragraph{Inferences of LVLMs.} We set the temperature as 0 or set the do sample as false to ensure reproducibility. For GPT-4o, we employ the official API. For InternVL V3, we employ it using two RTX4090 GPUs with 24GB of memory.
\paragraph{Trainable Detectors} To obtain a fair comparison, we set the hyperparameters the same for each detector in each fold. Every baseline can be held in one RTX4090 GPU with 24GB of memory. Meanwhile, we run each baseline five times and report the run with the best micro f1-score for each fold. Table~\ref{tab: hyperparameters} presents the hyperparameters of each baseline. We also provide the related codes in the supplementary material.

\section{Ablation Study}
We present the complete ablation study in Table~\ref{tab: ablation}. 

\begin{figure}[t]
    \centering
    \includegraphics[width=\linewidth]{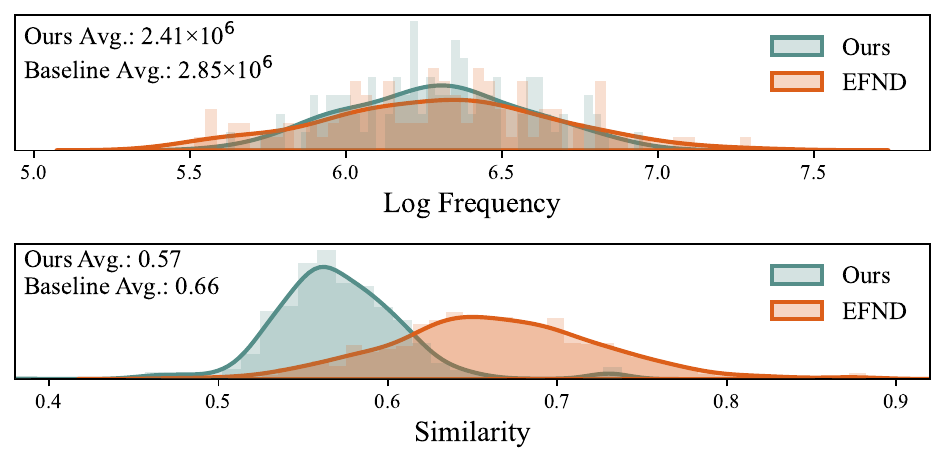}
    \caption{Token frequency and similarity distributions of \ours{} and a EFND. \ours{} presents a lower frequency and similarity, proving the diversity of the generated rationales.}
    \label{fig: diversity}
\end{figure}

\section{Human Evaluation}
\paragraph{Evaluation Guideline Document.} We first provide a brief guideline document for each expert. The guideline content is as follows:

Large vision language models are proven helpful in enhancing multimodal misinformation detection. 

A widely used paradigm, LVLM-as-enhancer, proposes to generate external explanations/rationales to enhance the performance of trainable detectors.

However, the explanations/rationales generated by LVLMs suffer from three challenges:
\begin{itemize}
    \item Lack of diversity: the rationales do not provide multiple perspectives for analyzing the news articles
    \item Lack of factuality: the rationales might contain factual errors
    \item Lack of relevance: the rationales might contain noisy information that are not helpful for judging.
\end{itemize}

Thus, this evaluation aims to evaluate which rationale is better to help judge the veracity of a specific news article.

Files in 'news\_articles' contain the textual content of a specific news article and two corresponding rationales.

Files in 'images' contain the visual image of a news article.

You need to enter your preferred rationale (based on diversity, factuality, and relevance) in 'answer.csv' for each news article.
\begin{itemize}
    \item 1 for preference of explanation 1
    \item 2 for preference of explanation 2
    \item 3 for no clear preference
\end{itemize}

Please follow your subjective feelings.

\paragraph{Major Voting} Each human evaluator needs to evaluate 200 rationale pairs, where the selected pairs are the same for all evaluators. For each pair, we employ major voting to obtain the final results. Notably, if the three experts have distinct answers, we consider the final result to be ``indistinguishable''.

\section{Diversity Analysis}
Figure~\ref{fig: diversity} shows the distributions of token frequency and similarity.

\end{document}